\documentclass[11pt,a4paper]{article}
\usepackage[hyperref]{emnlp2020}
\usepackage{times}
\usepackage{latexsym}

\usepackage{amsmath,amssymb}
\usepackage{enumitem}
\usepackage{color,soul}
\usepackage{multirow}
\usepackage{graphicx}

\usepackage{microtype}

\aclfinalcopy % Uncomment this line for the final submission
\title{An information theoretic view on selecting linguistic probes}

\author{Zining Zhu$^{1,2}$, Frank Rudzicz$^{3,4,1,2}$ \\
$^1$ University of Toronto, $^2$ Vector Institute, $^3$ Surgical Safety Technologies \\
$^4$ Li Ka Shing Knowledge Institute, St Michael's Hospital \\
\texttt{\{zining,frank\}@cs.toronto.edu}
}

\date{}

\begin{document}
\maketitle
\begin{abstract}
There is increasing interest in assessing the linguistic knowledge encoded in neural representations. A popular approach is to attach a diagnostic classifier -- or ``probe'' -- to perform supervised classification from internal representations. However, how to select a good probe is in debate. 
\citet{hewitt-liang-2019-designing} showed that a high performance on diagnostic classification itself is insufficient, because it can be attributed to either ``the representation being rich in knowledge'', or ``the probe learning the task'', which \citet{pimentel2020information} challenged. We show this dichotomy is valid information-theoretically.
In addition, we find that the methods to construct and select good probes proposed by the two papers, {\em control task} \citep{hewitt-liang-2019-designing} and {\em control function} \citep{pimentel2020information}, are equivalent -- the errors of their approaches are identical (modulo irrelevant terms). Empirically, these two selection criteria lead to results that highly agree with each other.
\end{abstract}

\section{Introduction}
Recently, neural networks have shown substantial progress in NLP tasks \citep{devlin2019bert,radford2019language}. To understand and explain their behavior, a natural question emerge: how much linguistic knowledge is encoded in these neural network systems? 

An efficient approach to reveal information encoded in internal representations uses diagnostic classifiers \citep{Alain2017}. Referred to as ``probes'', diagnostic classifiers are trained on pre-computed intermediate representations of neural NLP systems. The performance on tasks they are trained to predict are used to evaluate the richness of the linguistic representation in encoding the probed tasks. Such tasks include probing syntax  \citep{hewitt-manning-2019-structural-probe,lin2019open,tenney-etal-2019-bert}, semantics \citep{yaghoobzadeh-etal-2019-probing}, discourse features \citep{Chen2019,Liu2019,Tenney2019EdgeProbing}, and commonsense knowledge \citep{Petroni2019,Poerner2019}.

However, appropriate criteria for selecting a good probe is under debate. The traditional view that high-accuracy probes are better is challenged by \citet{hewitt-liang-2019-designing}, who proposed that the high accuracy could be attributed to either (1) that the representation contains rich linguistic knowledge, or (2) that the probe learns the task. To circumvent this ambiguity, they proposed to use the improvement of probing task performance against a control task (predicting random labels from the same representations), i.e., the ``selectivity'' criterion. 
Recently, \citet{pimentel2020information}, challenged this dichotomy from an information theoretic viewpoint. They proposed to use an ``information gain'' criterion, which empirically is the reduction in cross entropy from a ``control function task'' probing from randomized representation.

In this paper, we show the ``non-exclusive-or'' dichotomy raised by \citet{hewitt-liang-2019-designing} is valid information-theoretically. There is a difference between the original NLP model learning the task and the probe learning the task.

In addition, we show that the ``selectivity'' criterion and the ``control function'' criterion are comparably accurate. \citet{pimentel2020information} formulated their errors with the difference in a pair of KL divergences. We show that the error of the ``selectivity'' criterion \citep{hewitt-liang-2019-designing}, if measured from cross entropy loss, can be formulated in the difference in a pair of KL divergences as well. When randomizations are perfect, these two criteria differ by only constant terms.

Empirically, on a POS tag probing task on English, French and Spanish translations, we show that the ``selectivity'' and the ``control function'' criteria highly agree with each other. We rank experiments with over 10,000 different hyperparameter settings using these criteria. The Spearman correlation of the \citet{hewitt-liang-2019-designing} vs. \citet{pimentel2020information} criteria are on par with the correlations of ``accuracy vs. cross entropy loss'' -- two very strong baselines.

Overall, we recommend using control mechanisms to select probes, instead of relying on merely the probing task performance. When randomization is done well, controlling the target or representations are equivalent.

\section{Related work}
Diagnostic probes were originally intended to explain information encoded in intermediate representations \citep{adi2017fine,Alain2017,Belinkov2017}.
Recently, various probing tasks have queried the representations of, e.g., contextualized word embeddings \citep{tenney-etal-2019-bert,Tenney2019EdgeProbing} and sentence embeddings \citep{Linzen2016LSTMprobe,Chen2019,alt2020probing,Kassner2020negated,maudslay2020tale,Chi2020}.

The task of probing is usually formulated as a classifier problem, with the representations as input, and the features indicating information as output. A straightforward method to train a classifier is by minimizing cross entropy, which is the approach we follow. Note that \citet{voita2020information} derived training objectives from minimum description lengths, resulting in cross entropy loss and some variants.

%How to select good probes? 
%While cross entropy losses are widely adopted, two other criteria have recently been proposed for high quality probes \citep{hewitt-liang-2019-designing,pimentel2020information}. We show that these criteria are in fact equivalent, and that setting up controls for either the targets or the representations lead to probes with better information theoretic properties.

\section{Information theoretic probes}
\label{sec:infoprobe}
\subsection{Formulation}
We adopt the information theoretic formulation of linguistic probes of \citep{pimentel2020information}, and briefly summarize as follows.

We want to probe true labels $T$ from representations $R$. An ideal probe should accurately report the code-target mutual information $I(T;R)$, which is unfortunately intractable. We will write $I(T;R)$ in an alternative form.

Let $p(T\,|\,R)$ be the unknown true conditional distribution, and a diagnostic probe, according to the setting in literature \citep{Alain2017,hewitt-manning-2019-structural-probe,maudslay2020tale}, is an approximation $q_\theta (T\,|\,R)$ parameterized by $\theta$, then:

\begin{align*}
  I(T;R) &= H(T) - H(T\,|\,R) \\
  H(T\,|\,R) &= -\mathbb{E}_{p(T\,|\,R)} \text{ log } p(T\,|\,R) \\
    &= -\mathbb{E}_{p(T\,|\,R)} \text{ log }\frac{p(T\,|\,R) q_\theta (T\,|\,R)}{q_\theta (T\,|\,R)} \\
    &= -\mathbb{E}_{p} \text{ log }q_\theta - \mathbb{E}_{p} \text{ log} \frac{p}{q_\theta}\\
    &= H(p,q_\theta) - \text{KL}(p \,\|\, q_{\theta}),  
\end{align*}
where $p$ and $q_\theta$ stand for $p(T\,|\,R)$ and $q_\theta (T\,|\,R)$ respectively. We also use $H(p, q_\theta)=-\mathbb{E}_p \text{ log } q_\theta$ to represent the cross entropy for simplicity.

\subsection{The source of probing error}
\paragraph{A valid dichotomy}
Traditionally, people use the cross entropy loss of the diagnostic probe $H(p,q_\theta)$ to approximate $I(T;R)$.
We can derive a source of error by rewinding the above formulations:
$$H(p,q_\theta) = H(T)-I(T;R) + \text{KL}(p \,\|\, q_\theta)$$

The first term on RHS, $H(T)$, is independent of either $R$ or $\theta$. Therefore, a low cross entropy loss $H(p, q_\theta)$ can be caused by either of the two scenarios:
\begin{itemize}
\setlength{\itemsep}{0pt}
    \item High code-target mutual information $I(T;R)$, indicating the representation $R$ contains rich information about the target $T$.
    \item Low KL-divergence between $p(T\,|\,R)$ and $q_\theta(T\,|\,R)$, indicating the probe learns the task.
\end{itemize}

The two scenarios exactly correspond to the dichotomy of \citet{hewitt-liang-2019-designing}.

\paragraph{A good probe}
To get a good probe, we want $H_{q_\theta}(T\,|\,R)$ to approximate $I(T;R)$ as much as possible. This means a good probe should minimize $\text{KL}(p\,\|\, q_\theta)$, as proposed by \citet{pimentel2020information}.

However, empirically \citet{pimentel2020information} (as well as many previous articles) used the cross entropy loss $H_{q_\phi}(T\,|\,R)$ to select good probes, which is insufficient, as described above. Alternatively, \citet{hewitt-liang-2019-designing} and \citet{pimentel2020information} proposed control tasks and control functions, respectively.

\subsection{The control task}
The control task \citep{hewitt-liang-2019-designing} sets random targets for the probing task. Let us use $c(T)$ to indicate a ``control function'' applied on a token $v$ that originally has label $T$. The control function could nullify the information of the input, if necessary.

If we measure the difference between cross entropy in the control task and probing task $H(p(c(T)\,|\,R), q_{\theta_c}(c(T)\,|\,R))-H(p(T\,|\,R), q_\theta (T\,|\,R))$, we can derive a form of error margin in their measurements\footnote{Note that \citet{hewitt-liang-2019-designing} used accuracy instead of cross entropy. We discuss cross entropy so as to compare the errors against the control function \citep{pimentel2020information}}.
Let us use a short-hand notations $H(p_c, q_{\theta_c}) - H(p, q_\theta)$ for clarity. Now, what does the diff between cross entropy on control task and probing task actually contains? 
\begin{align*}
& H(p_c, q_{\theta_c}) - H(p, q_\theta) \\
=& \left(H(c(T)) - I(c(T);R) + \text{KL}(p_c\,\|\, q_{\theta_c}) \right) \\
 &- \left( H(T) - I(T;R) + \text{KL}(p\,\|\, q_\theta)\right) \\
=& \left(  H(c(T)) - H(T)\right) \\
&- I(c(T);R) + I(T;R) \\
&+  \left(\text{KL}(p_c\,\|\, q_{\theta_c}) - \text{KL}(p\,\|\, q_\theta) \right)
\end{align*}
We already knew that $H(T)=\text{Const}$. According the definition of control function, the output $c(T)$ would be independent of $R$. Then:
\begin{align*}
  & H(c(T))=\text{Const}  \\
  & p(c(T),R)=p(c(T))p(R) \\
  & I(c(T);R)=\mathbb{E} \frac{p(c(T), R)}{p(c(T))p(R)} = \text{Const}
\end{align*}

Therefore:
$$ H(p_c, q_{\theta_c}) - H(p, q_\theta) = I(T;R) - \Delta_h$$
where $\Delta_h$ is a short-hand notation for the measurement error in the \emph{control task} criteria:
\begin{align}
\Delta_{h}=\text{KL}(p\,\|\, q_\theta ) - \text{KL}(p_c\,\|\,q_{\theta_c}) + \text{Const}
\label{eq:delta_h}
\end{align}
When the probe fits the true distribution to a similar extent on both the control task and probing task, the error $\Delta_h$ would be small. Unfortunately, both KL terms are intractable. 

\subsection{The control function}
Control function \citep{pimentel2020information} introduces a random processor $\mathbf{c}(\cdot)$ on the representation $R$. To measure the information gain, they used an ``information gain'' criterion:
$$\mathcal{G}(T,R,\mathbf{c}) = I(T;R) - I(T;\mathbf{c}(R))$$
Noticing that mutual information terms are intractable, they approximated the objective with the difference between cross entropy in the control function task (we refer to as ``control function'' henceforth) and the probing task: 
\begin{align*}
    \tilde{\mathcal{G}}(T,R,\mathbf{c})=H(p_c,q_{\phi_c}) - H(p, q_\phi)
\end{align*}
To compute the error of this approximation, they reformulated the terms as following:
\begin{align*}
    \mathcal{G}&(T,R,\mathbf{c}) \\
    &= H(T)-H(T\,|\,R) - H(T)+H(T\,|\,\mathbf{c}(R))
\end{align*}
The two $H(T)$ terms cancel out, then:
\begin{align*}
H(T\,|\,R) =& H(p(T\,|\,R),q_\phi (T\,|\,R)) \\
 &- \text{KL}(p(T\,|\,R)\,\|\, q_\phi (T\,|\,R)) \\
 &= H(p,q_\phi) - \text{KL}(p, q_\phi)\\
H(T|\mathbf{c}(R)) =& H(p(T\,|\,\mathbf{c}(R)),q_{\phi_c}(T\,|\,\mathbf{c}(R))) \\
 &- \text{KL}(p(T\,|\,\mathbf{c}(R))\,\|\, q_{\phi_c}(T\,|\,\mathbf{c}(R))) \\
 &= H(p_c, q_{\phi_c}) - \text{KL}(p_c\,\|\, q_{\phi_c})
\end{align*}
Where we abbreviate similarly as we did for the control task. Specifically, we write $\phi$ for the probe parameters of control function to tell apart from $\theta$ in the control task.

\citet{pimentel2020information} showed that the error of their approximation, $\Delta_p = \mathcal{G}(T,R,\mathbf{c}) - \tilde{\mathcal{G}}(T, R, \mathbf{c})$, can be expressed as:
\begin{align}
\Delta_p = \text{KL}(p\,\|\, q_\phi) - \text{KL}(p_c\,\|\, q_{\phi_c})
\label{eq:delta_p}
\end{align}

Again, when the probe fits the true distribution to a similar extent on both the target labels distribution and the probing task, the $\Delta_p$ will be small. Unfortunately, both KL terms are intractable too.

% Moved here for better formatting
\begin{table*}
    \centering
    \begin{tabular}{c c c c c c}
        \hline 
        \multirow{2}{*}{Language} & \multirow{2}{*}{\# POS} & \textbf{\# Tokens} &  \multicolumn{3}{c}{\textbf{Correlations}} \\
        & & train / dev / test & (\textbf{t\_acc},\textbf{f\_ent}) & (\textbf{t\_acc},\textbf{t\_ent}) & (\textbf{f\_acc},\textbf{f\_ent}) \\ \hline 
        English & 17 & 177k / 22k / 22k & 0.1615 & 0.1334 & 0.1763  \\  
        French & 15 & 303k / 31k / 8k  & 0.0906 & 0.0606 & 0.1295 \\ 
        Spanish & 16 & 341k / 33k / 11k & 0.1360 & 0.0560 & 0.1254 \\ \hline 
    \end{tabular}
    \caption{Spearman correlations between t\_acc (the ``selectivity'' criterion \citep{hewitt-liang-2019-designing}) and f\_ent (the ``gain'' criterion \citep{pimentel2020information}) are on par with two ``accuracy vs. cross entropy'' correlations.}
    \label{tab:results_table}
\end{table*}

\subsection{Control task vs control function}
From Equations \ref{eq:delta_h} and \ref{eq:delta_p}, we showed that the \emph{selectivity} criterion of \citet{hewitt-liang-2019-designing} and the \emph{information gain} criterion of \citet{pimentel2020information}, if both measured in cross entropy loss, have very similar errors in approximating information gains. 

These errors, $\Delta_h$ and $\Delta_p$ respectively, appear in very similar forms\footnote{When the two $c(\cdot)$ are ideal, $\Delta_h$ and $\Delta_p$ differ by only irrelevant terms -- we include the derivations in Supplementary Material.}.
The probes selected from these two criteria should be highly correlated to each other, and our experiments will confirm.

\section{Experiments}

\subsection{Setup}
We use the same family of probes as \citet{hewitt-liang-2019-designing} and \citet{pimentel2020information}, multiple layers perceptrons with ReLU activations, to show the correlations of their ``good probe'' criteria (control task and control function, respectively). 

Overall, we sweep the probe model hyper-parameters with a unified training scheme on three tasks (probe, control task, control function). The control task (function) setting includes labels (embeddings) drawn from a uniform random sample once before all experiments.
In each training, we follow the setting of \citep{hewitt-liang-2019-designing}. We save the model with the best dev loss, report the test set loss and accuracy, and average across 4 different random seeds.

\paragraph{Data}
We use the Universal Dependency \citep{UniversalDependencies2.5} dataset loaded with the Flair toolkit \citep{akbik2018flair}. We examine three languages: English, French, and Spanish. 
For the probing task, we use POS with labels provided by SpaCy\footnote{\url{https://spacy.io}}. 
We use the embedding of multilingual BERT (mBERT) implemented by huggingface \citep{HuggingfaceTransformers2019}. If a word is split into multiple word pieces, we average its representations.

\subsection{The ``good probes'' are good for both}
When measuring the qualities of probes using the ``selectivity'' \citep{hewitt-liang-2019-designing} or ``information gain'' \citep{pimentel2020information} criterion, we show that the rules-of-thumb for training good probes largely agree.
\begin{itemize}[nosep]
    \item Early stopping before 24,000 gradient steps (approximately 4 epochs) could inhibit probe quality, but longer training procedures do not improve the probe qualities considerably.
    \item Smaller probes are better in general, but exceptions exist. For example, when weight decay is set to 0, probes with one hidden layer and 40 hidden neurons are better in both criteria.
    \item A small weight decay is beneficial.
\end{itemize}
We include more descriptions, including comprehensive experiment configurations and plots in the Supplementary Material.

\subsection{The high correlation between criteria}
In addition to the qualitative correlations shown above, we compute the correlations between the two criteria over a grid-search style hyper-parameter sweep of over 10,000 configurations.
For each ``probe, control task, control function'' experiment set, we record the following four criteria:
\begin{itemize}[nosep]
    \item \textbf{t\_acc}: Difference between probing task and control \textul{\textbf{t}}ask \textul{\textbf{acc}}uracy. This is the ``selectivity'' criterion of \citet{hewitt-liang-2019-designing}. 
    \item \textbf{f\_ent}: Difference between control \textul{\textbf{f}}unction and probing task cross \textul{\textbf{ent}}ropy. This is the ``gain'' criterion of \citet{pimentel2020information}.
    \item \textbf{t\_ent}: Difference between the control \textbf{\textul{t}}ask and the probing task cross \textul{\textbf{ent}}ropy.
    \item \textbf{f\_acc}: Difference between the probing task and control \textul{\textbf{f}}unction \textul{\textbf{acc}}uracy.
\end{itemize}
We collect all experiments of each language according to these criteria, and use Spearman correlation to test three pairs of correlations. As is reported in Table \ref{tab:results_table}, the (t\_acc, f\_ent) correlations are comparable to two strong baselines, (t\_acc, t\_ent) and (f\_acc, f\_ent), the correlations between measurements in accuracy and cross entropy losses.

\section{Conclusion}
When selecting probes that better approximate $I(T;R)$, we recommend measuring \emph{with a control mechanism} instead of relying on the traditional cross entropy on probing task. We show both information-theoretically and empirically, that controlling the targets and representations are equivalent, as long as the control mechanism is randomized.

\section{Acknowledgement}
We would like to thank Mohamed Abdalla for his insights and discussion. Rudzicz is supported by a CIFAR Chair in artificial intelligence.

\bibliography{emnlp2020}
\bibliographystyle{acl_natbib}

\newpage
\pagebreak
\appendix

\section{Difference between the two criteria}
In Section \ref{sec:infoprobe} we show that the error of the two criteria can both be written as difference between a pair of KL divergence (modulo a constant term). Here we further simplify the terms when we assume the control task and functions take perfectly random distributions (i.e., independent of the task and representations, respectively).
\begin{align*}
&\Delta_h - \Delta_p = \text{Const} + \\
 &\text{KL}(p(c(T)|R) \,\|\, q_{\theta_c}(c(T)|R)) - \text{KL}(p\,\|\, q_\theta) \\
 &- \text{KL}(p(T| c(R)) \,\|\, q_{\phi_c}(T| c(R))) + \text{KL}(p\,\|\, q_\phi) \\
\end{align*}
When we use the same hyperparameter setting, $q_\theta(T\,|\,R)$ and $q_\phi (T\,|\,R)$ should be able to approximate $p(T\,|\,R)$ to the same extent, so $\text{KL}(p\,\|\, q_\theta)$ and $\text{KL}(p\,\|\, q_\phi)$ cancel out. Additionally, following the definitions of the control function and control tasks, we can simplify as follows:
\begin{align*}
    &p(c(T)|R) = p(c(T)), q_{\theta_c}(c(T)|R) = q_{\theta_c} (c(T)) \\
    &p(T|c(R)) = p(T), q_{\phi_c}(T|c(R)) = q_{\phi_c}(T) 
\end{align*}
\begin{align*}
\displaystyle
    \text{KL}&(p(c(T)|R) \,\|\, q_{\theta_c}(c(T)|R))= \mathbb{E}_{p(c(T))} \frac{p(c(T))}{q_{\theta_c}(c(T))} \\
    \text{KL}&(p(T| c(R)) \,\|\, q_{\phi_c}(T| c(R))) = \mathbb{E}_{p(T)} \frac{p(T)}{q_{\phi_c}(T)}
\end{align*}

Therefore, the difference between errors of criteria of \citet{hewitt-liang-2019-designing} and \citet{pimentel2020information} are:
\begin{align}
\begin{split}
\Delta_h - \Delta_p &= \text{Const} - \text{KL}(p(T) \,\|\, q_{\phi_c}(T)) \\
&+ \text{KL}(p(c(T)) \,\|\, q_{\theta_c}(c(T))) 
\label{eq:final_diff}
\end{split}
\end{align}

In short, these two criteria differ by terms dependent only on the randomization functions and the inherent distributions of task labels, i.e., irrelevant terms.

\section{Experiments details}
\citet{hewitt-liang-2019-designing} proposed some rules-of-thumb to select good probes, including a ``simple probe'' suggestion. We sweep the hyper parameters to test whether these rules also apply when measuring probe qualities using the control function \citep{pimentel2020information}.

\begin{figure*}
    \centering
    \includegraphics[width=\linewidth]{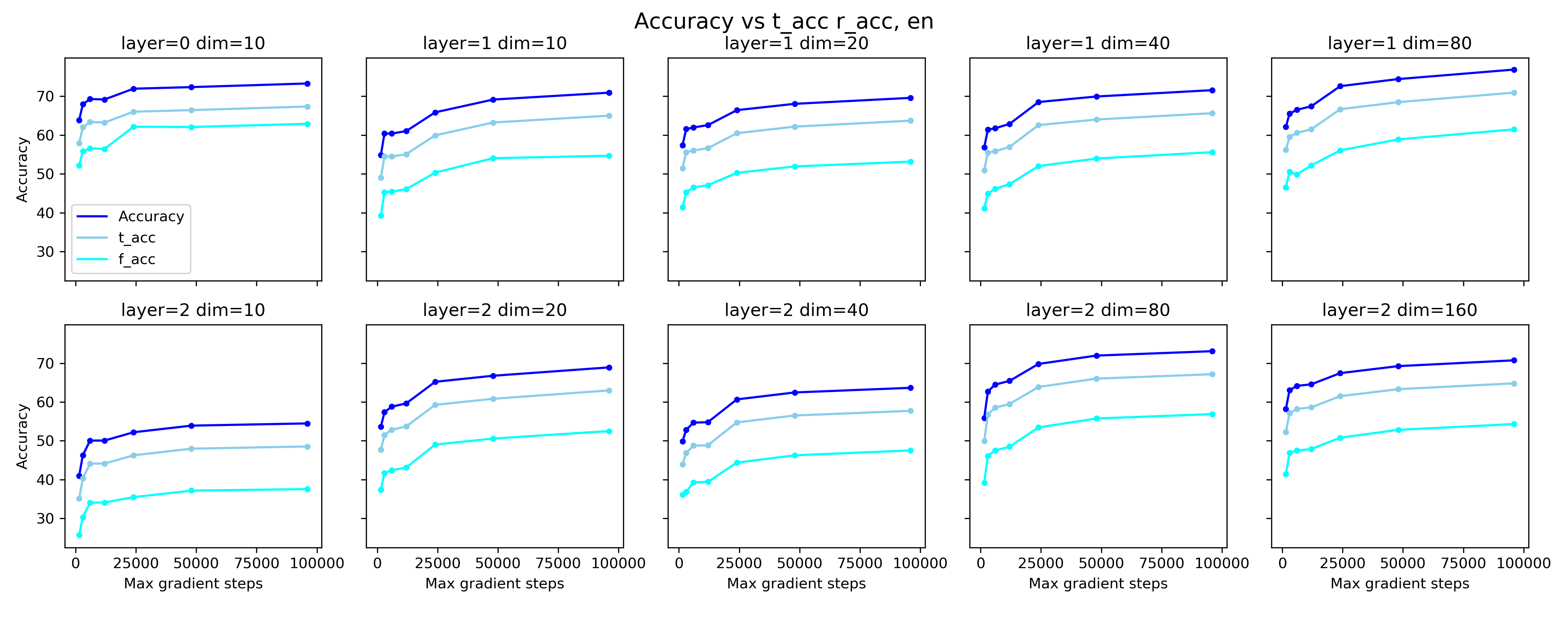}
    \includegraphics[width=\linewidth]{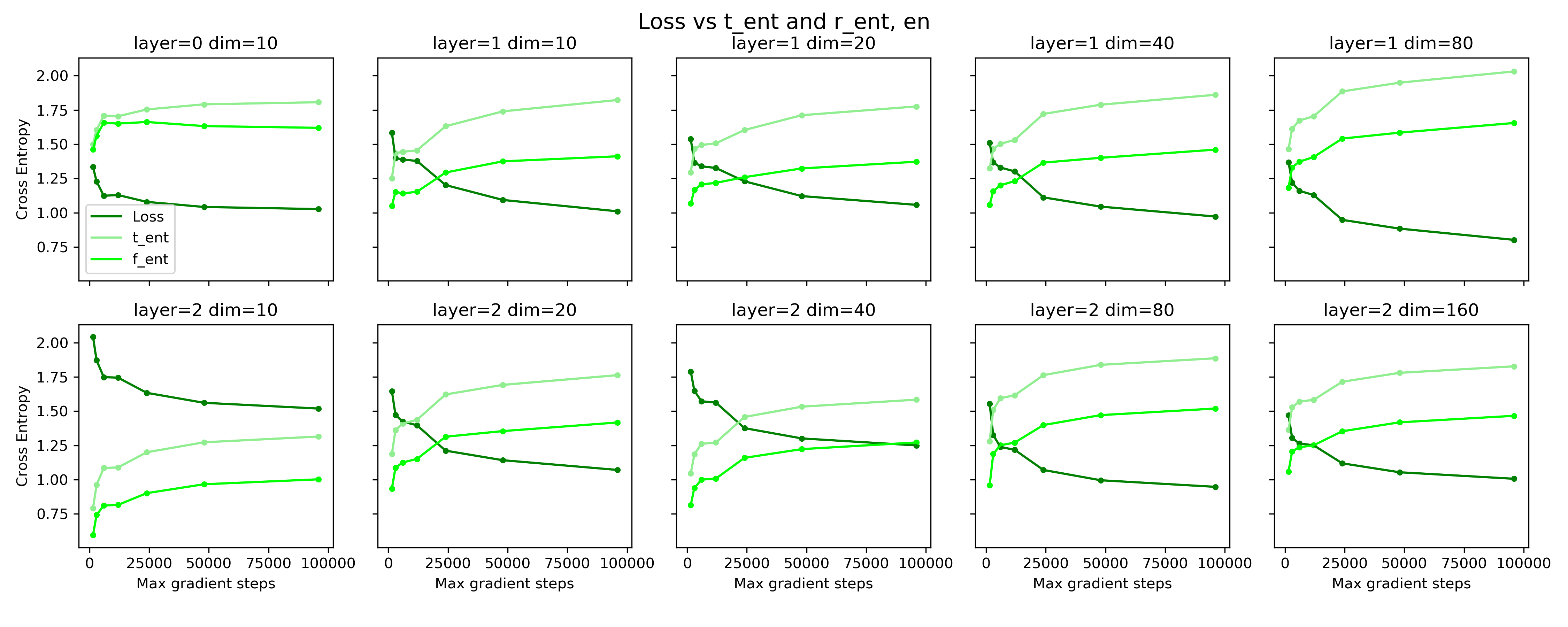}
    \caption{Max gradient step vs accuracy, t\_acc, f\_acc (blue) and loss, t\_ent, f\_ent (green) on English. The ``t'' refers to control task \citep{hewitt-liang-2019-designing}, and ``h'' refers to control function \citep{pimentel2020information}. In these set of experiments, we look for the best learning rate and zero weight decay in each configuration.}
    \label{fig:early_stopping_en}
\end{figure*}

\begin{figure*}
    \centering
    \includegraphics[width=.49\linewidth]{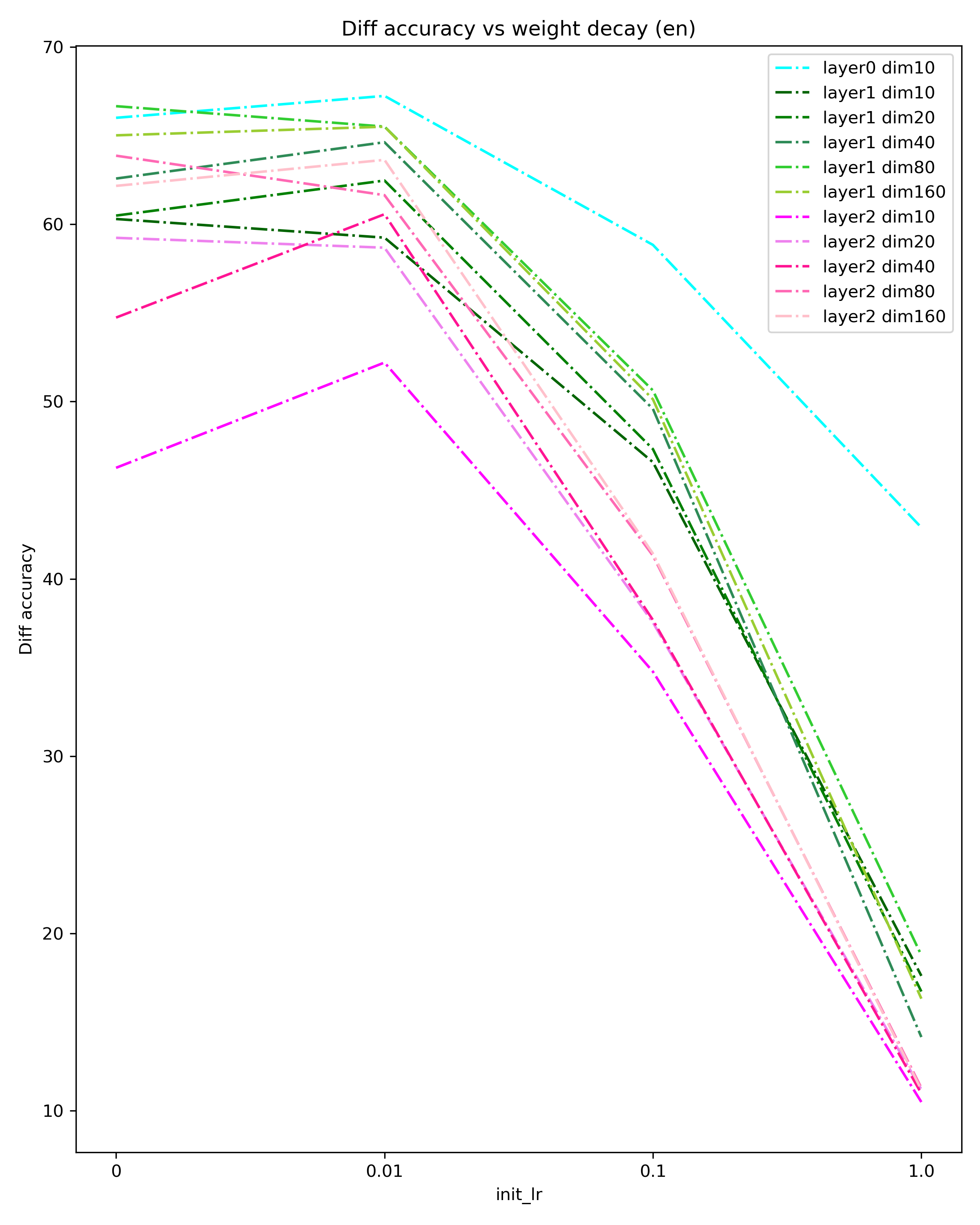}
    \includegraphics[width=.49\linewidth]{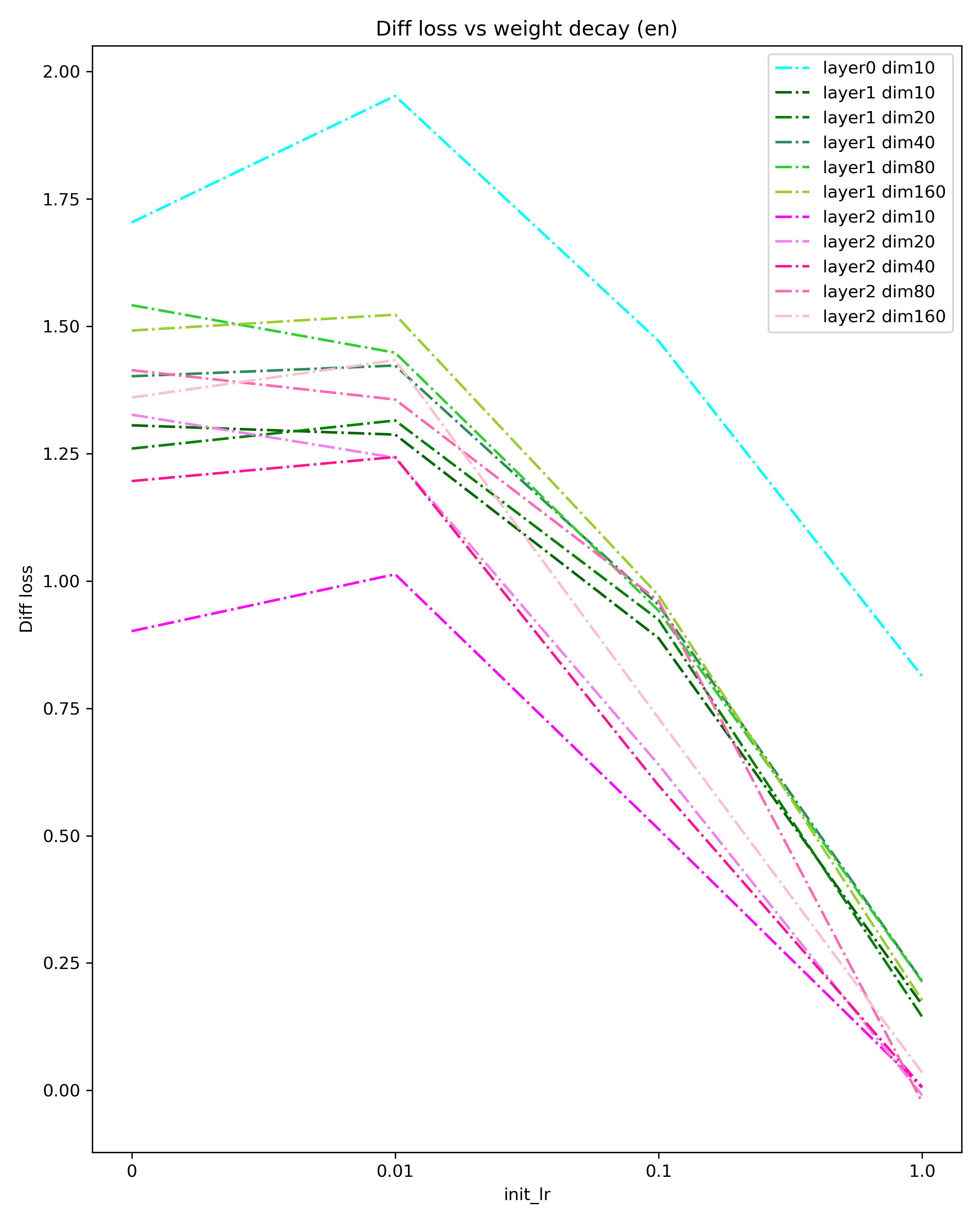}
    \caption{The ``difference of accuracy'' \citep{hewitt-liang-2019-designing} and the ``difference of loss'' \citep{pimentel2020information} criteria against weight decay on model configurations, on UD English. For each configuration, the learning rate leading to the highest accuracy is selected.}
    \label{fig:weight_decay_vs_criteria_en}
\end{figure*}

\begin{figure*}
    \centering
    \includegraphics[width=.49\linewidth]{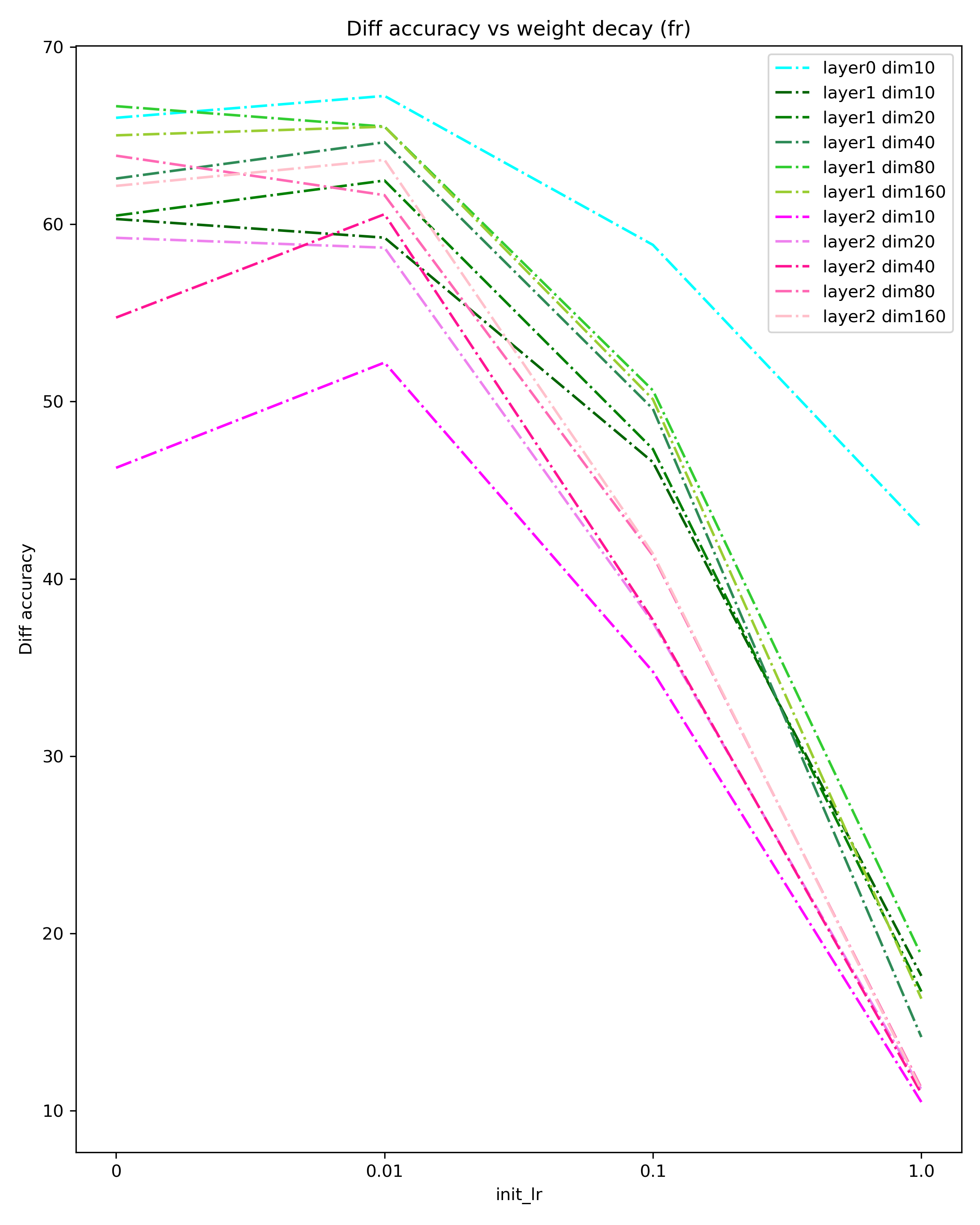}
    \includegraphics[width=.49\linewidth]{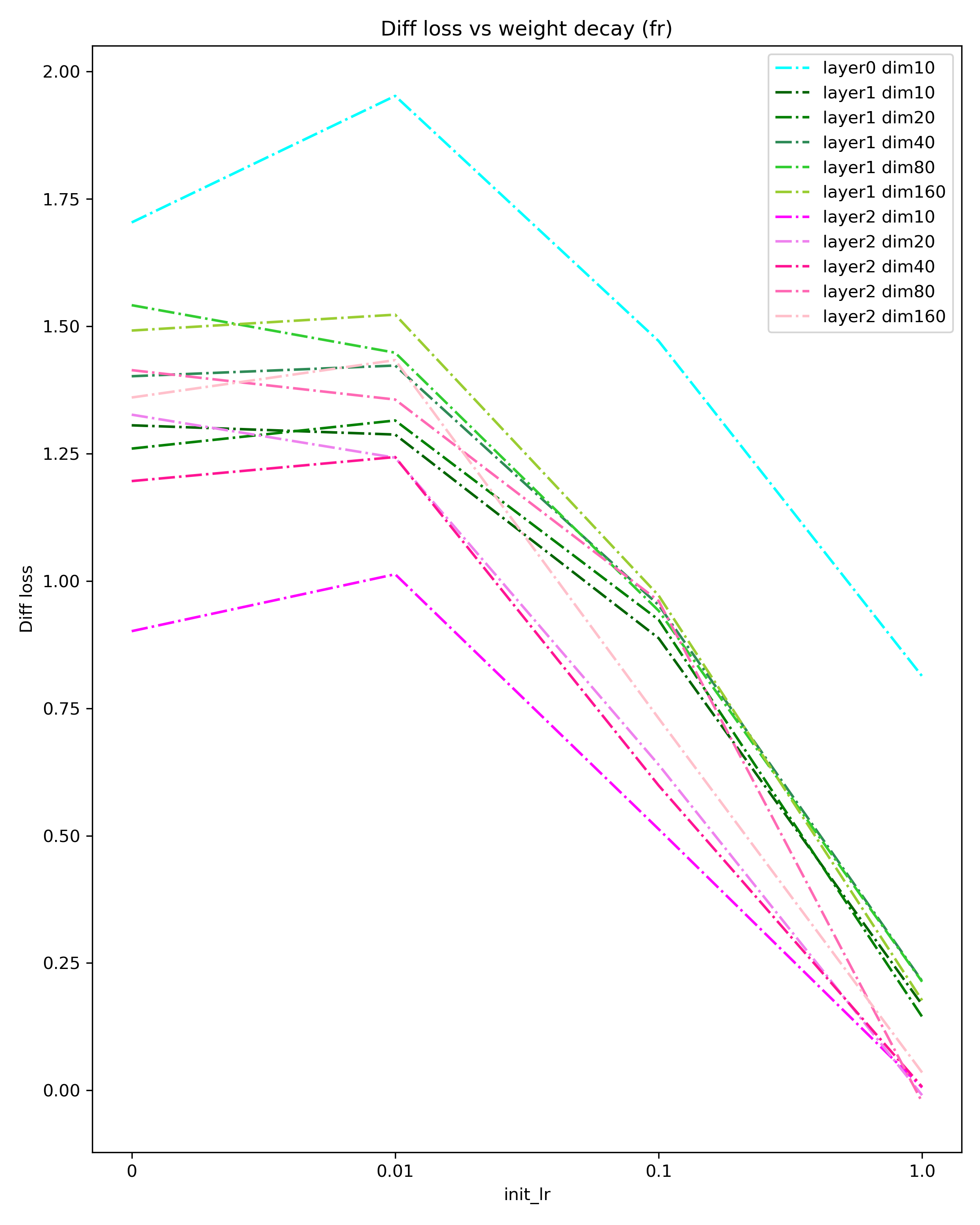}
    \caption{The ``difference of accuracy'' \citep{hewitt-liang-2019-designing} and the ``difference of loss'' \citep{pimentel2020information} criteria against weight decay on model configurations, on UD French. For each configuration, the learning rate leading to the highest accuracy is selected.}
    \label{fig:weight_decay_vs_criteria_fr}
\end{figure*}

\begin{figure*}
    \centering
    \includegraphics[width=.49\linewidth]{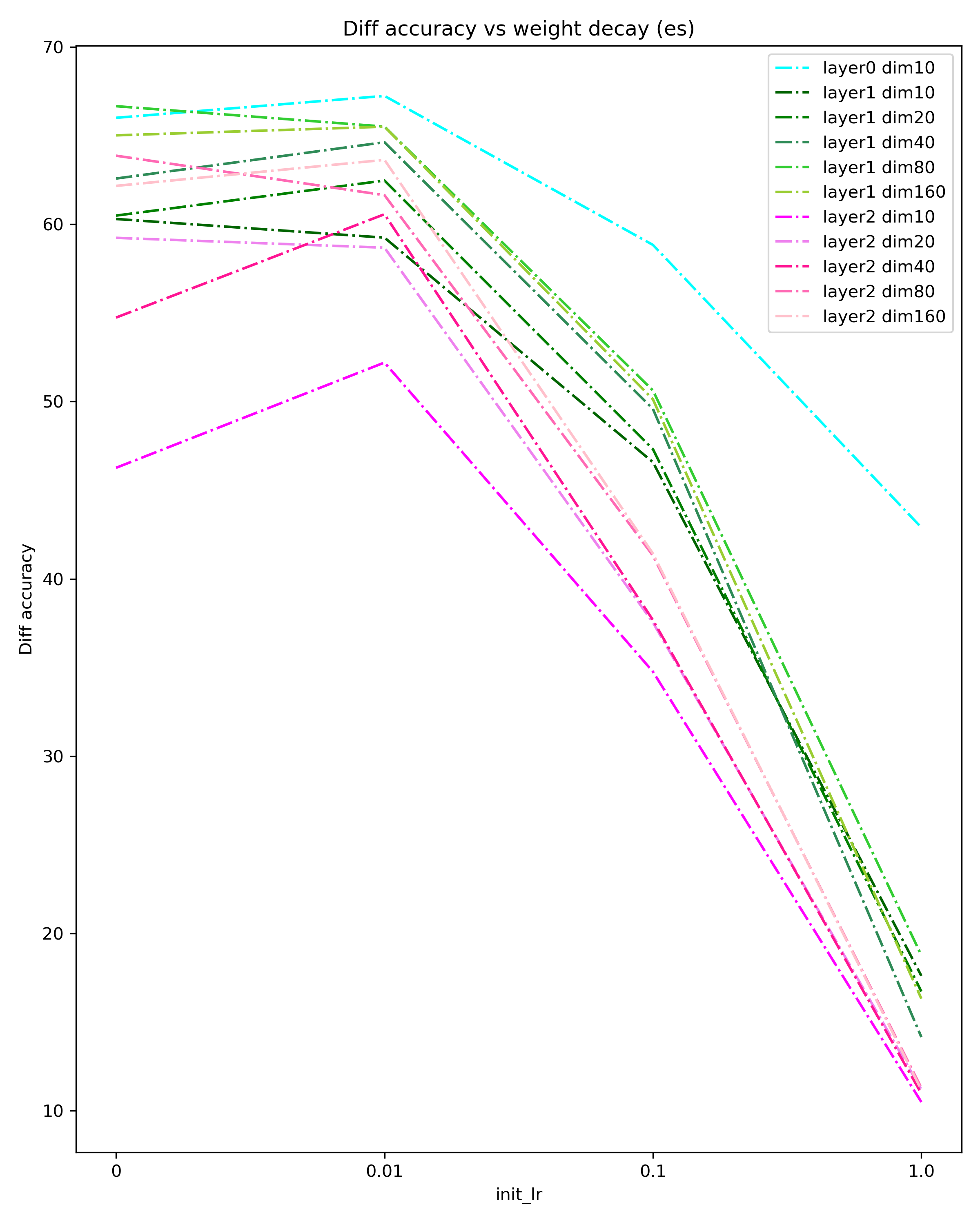}
    \includegraphics[width=.49\linewidth]{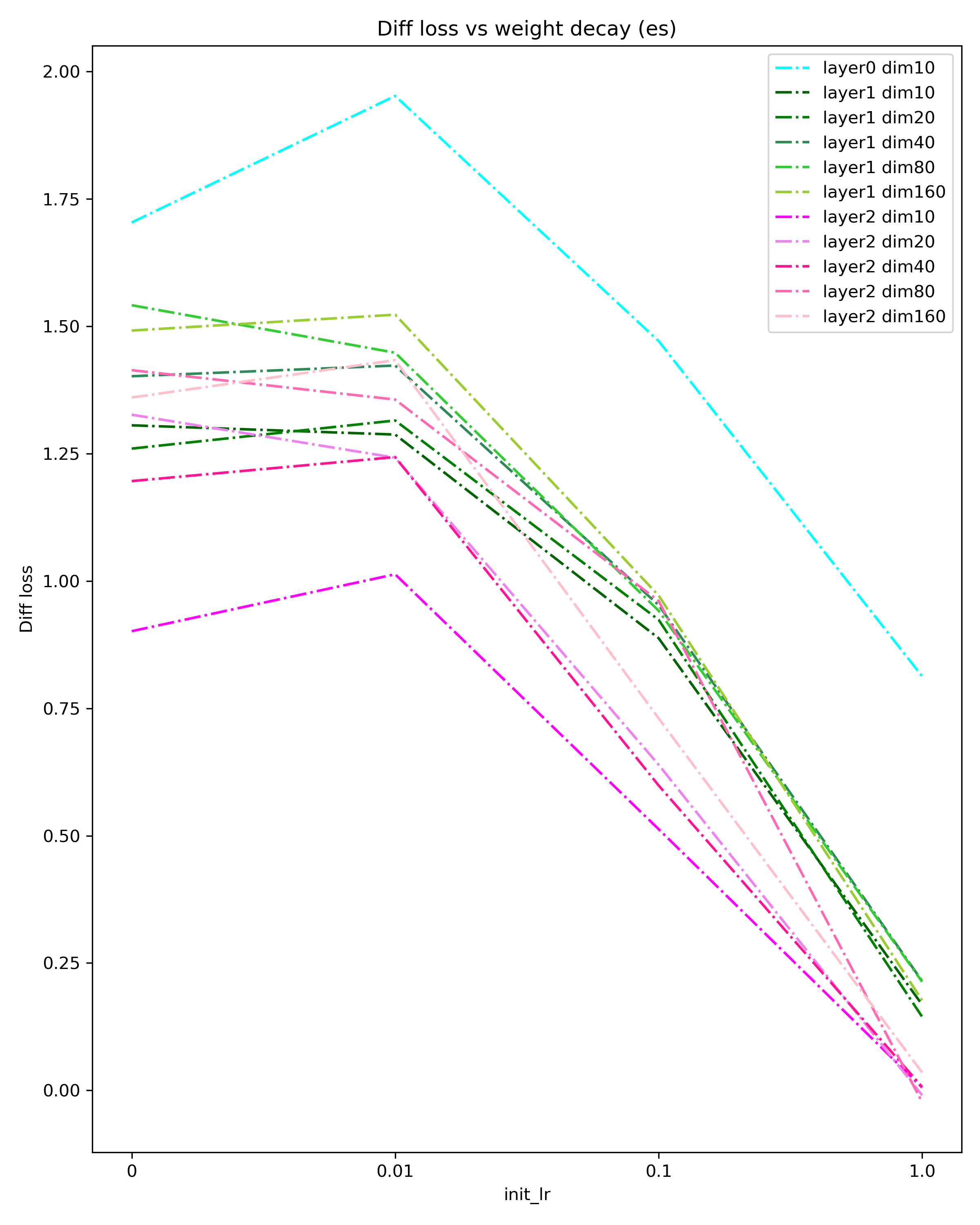}
    \caption{The ``difference of accuracy'' \citep{hewitt-liang-2019-designing} and the ``difference of loss'' \citep{pimentel2020information} criteria against weight decay on model configurations, on UD Spanish. For each configuration, the learning rate leading to the highest accuracy is selected.}
    \label{fig:weight_decay_vs_criteria_es}
\end{figure*}

\paragraph{Hyper-parameter ranges}
We sweep hyper parameters from the following ranges:
\begin{itemize}[nosep]
    \item Learning rate: $\{10^{-4}, 5\times 10^{-5}, 3\times 10^{-5}, 10^{-5}, 5\times 10^{-6}, 3\times10^{-6}\}$
    \item Maximum gradient steps: $\{1500, 3000, 6000, 12000, 24000, 96000, \infty\}$. Their effects are shown in Figure \ref{fig:early_stopping_en}. 
    \item Weight decay: $\{ 0, 0.01, 0.1, 1.0 \}$. Their effects are shown in Figures \ref{fig:weight_decay_vs_criteria_en}, \ref{fig:weight_decay_vs_criteria_fr} and \ref{fig:weight_decay_vs_criteria_es}. When sweeping weight decay, max gradient step is set to 24000.
\end{itemize}
In any configuration mentioned above, we run four experiments with random seeds 73, 421, 9973, 361091, and average the reported results (i.e., accuracy and loss).

\paragraph{Early stopping could inhibit probe quality}
Early stopping, if stopped before 24,000 gradient steps (approximately 4 epochs) may inhibit the quality of probes. In addition, we Figure \ref{fig:early_stopping_en} shows high correlation between the ``selectivity'' \citep{hewitt-liang-2019-designing} and ``information gain'' \citep{pimentel2020information} criteria.

\paragraph{Small weight decay is beneficial}
We find that smaller weight decays (e.g., 0.01) are more beneficial for probes than larger weight decays. While the two criteria rank the capacity of probes similarly, the most simple probes tend to stand out more distinctively with the ``selectivity'' criterion \citep{pimentel2020information}, as are shown on Figures \ref{fig:weight_decay_vs_criteria_en}, \ref{fig:weight_decay_vs_criteria_fr}, and \ref{fig:weight_decay_vs_criteria_es}.

\begin{figure*}
    \centering
    \includegraphics[width=.49\linewidth]{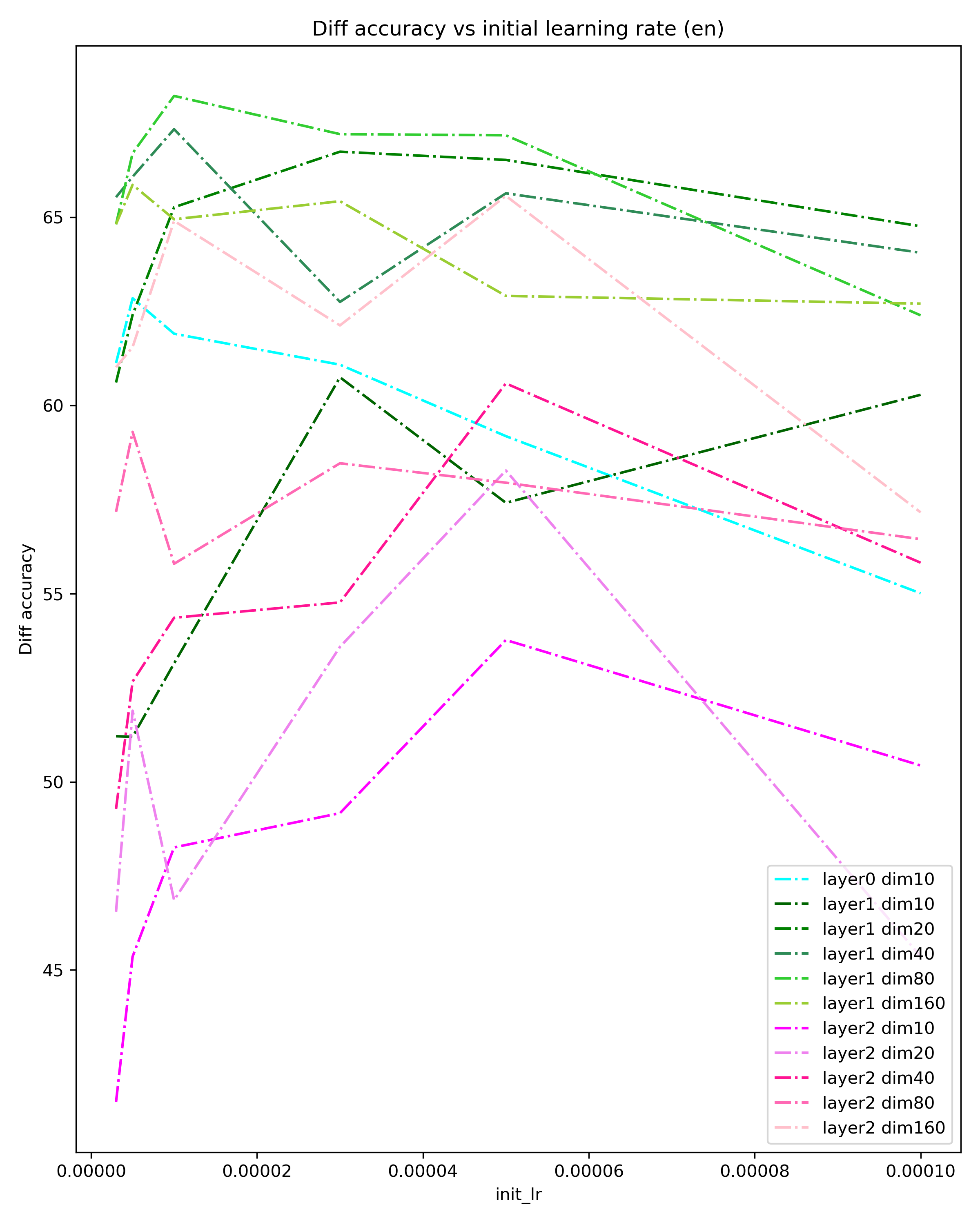}
    \includegraphics[width=.49\linewidth]{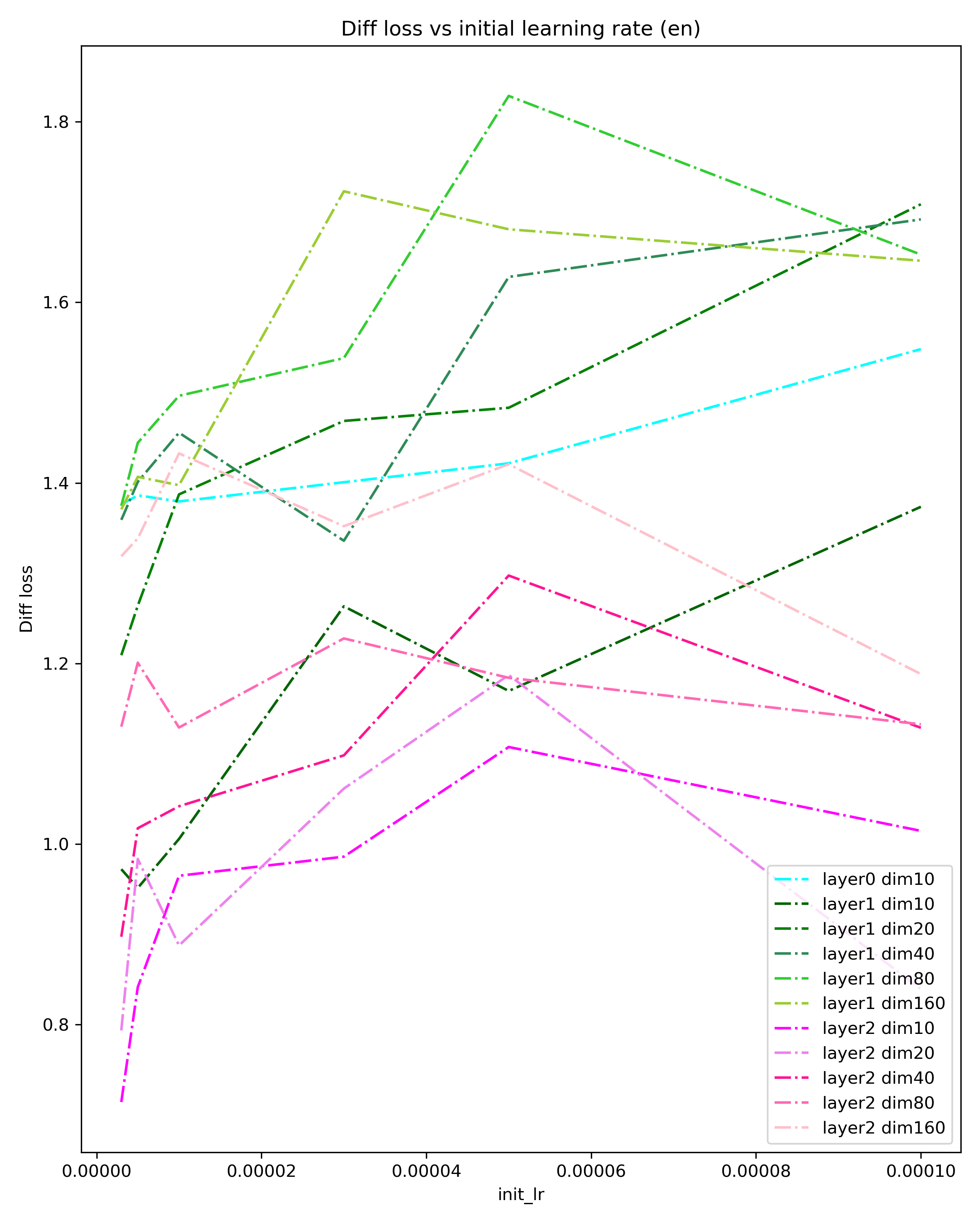}
    \caption{The ``difference of accuracy'' \citep{hewitt-liang-2019-designing} and the ``difference of loss'' \citep{pimentel2020information} criteria with different learning rates on model configurations, on UD English. The weight decay is set to 0.}
    \label{fig:lr_vs_criteria_en}
\end{figure*}

\begin{figure*}
    \centering
    \includegraphics[width=.49\linewidth]{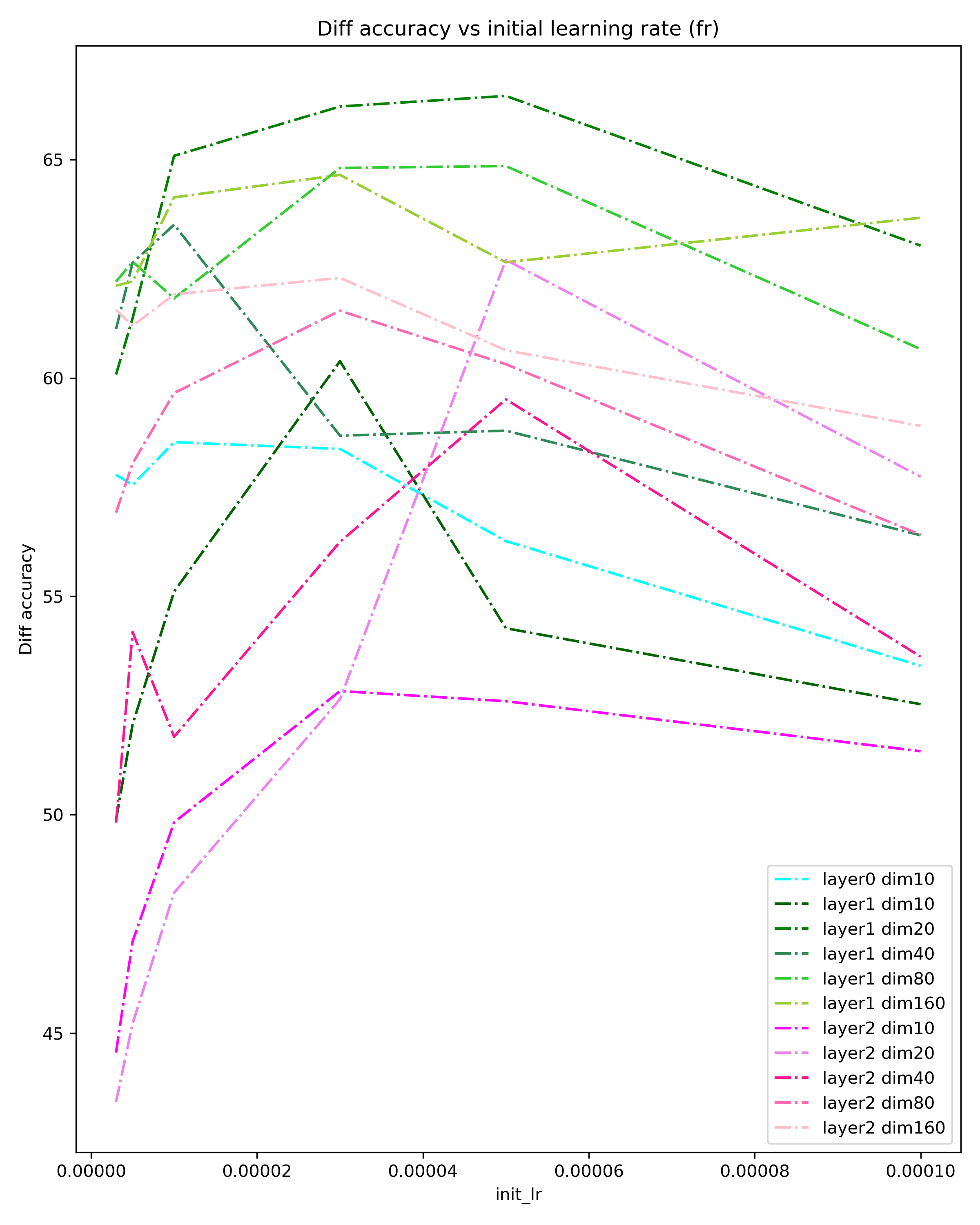}
    \includegraphics[width=.49\linewidth]{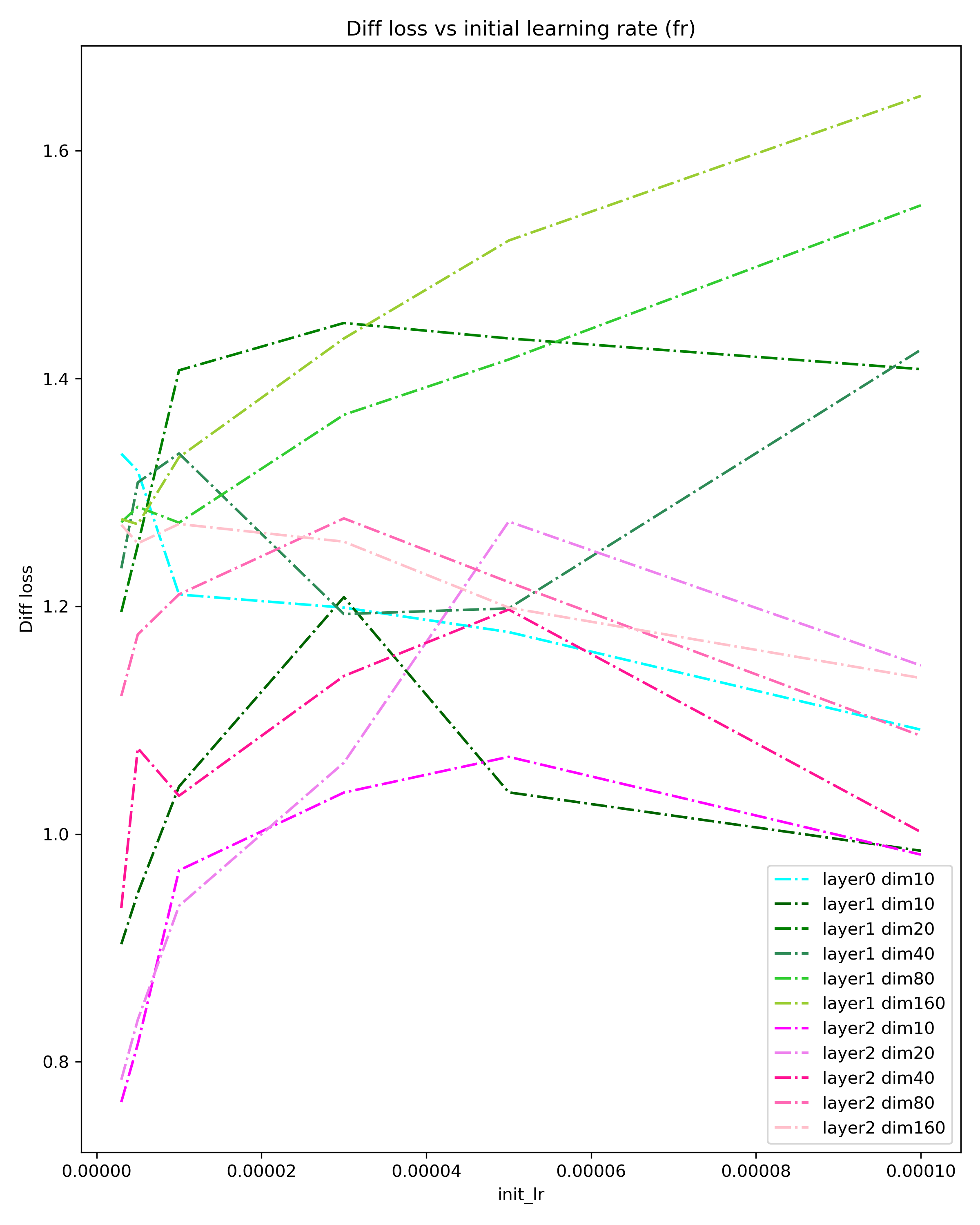}
    \caption{The ``difference of accuracy'' \citep{hewitt-liang-2019-designing} and the ``difference of loss'' \citep{pimentel2020information} criteria with different learning rates on model configurations, on UD French. The weight decay is set to 0.}
    \label{fig:lr_vs_criteria_fr}
\end{figure*}

\begin{figure*}
    \centering
    \includegraphics[width=.49\linewidth]{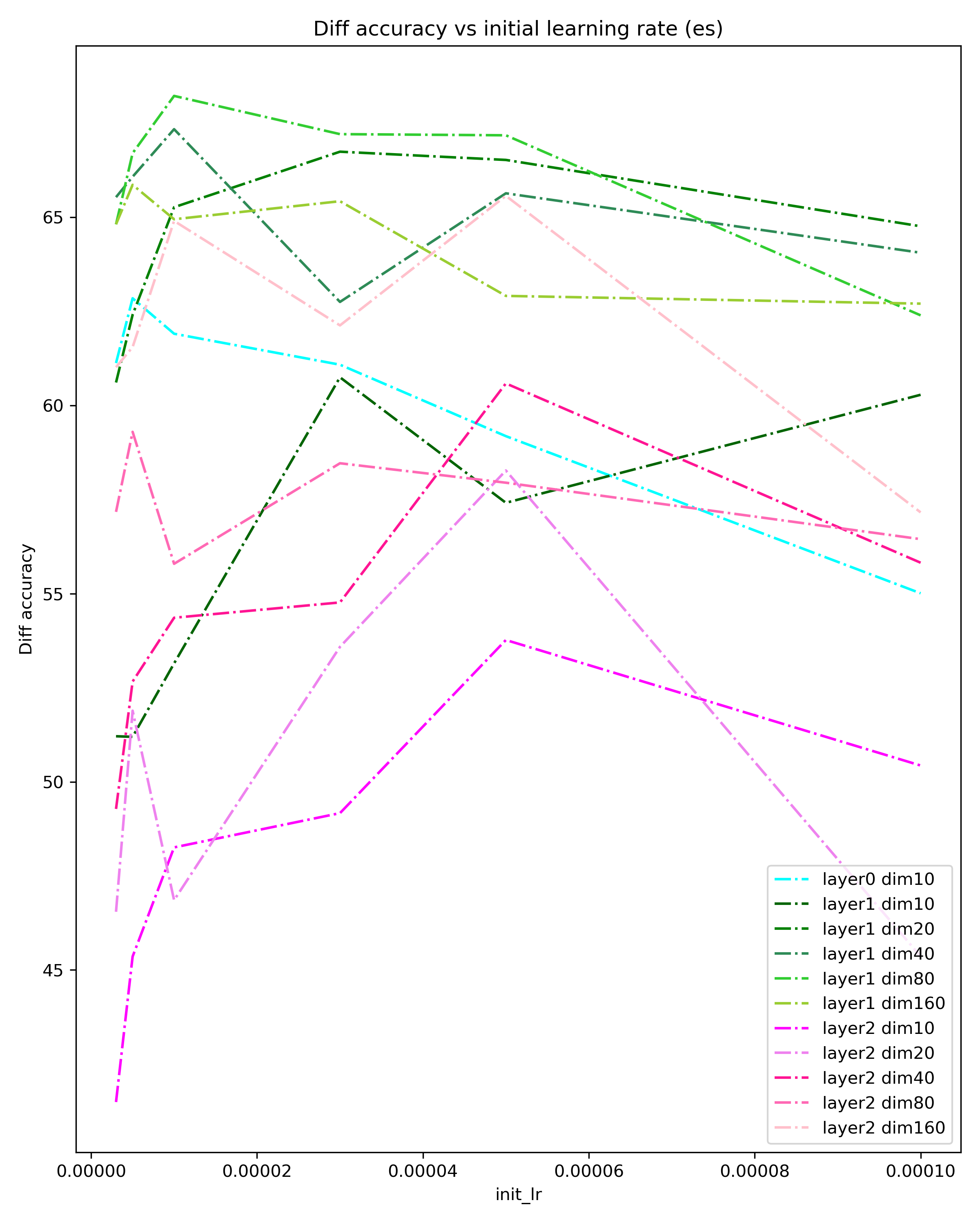}
    \includegraphics[width=.49\linewidth]{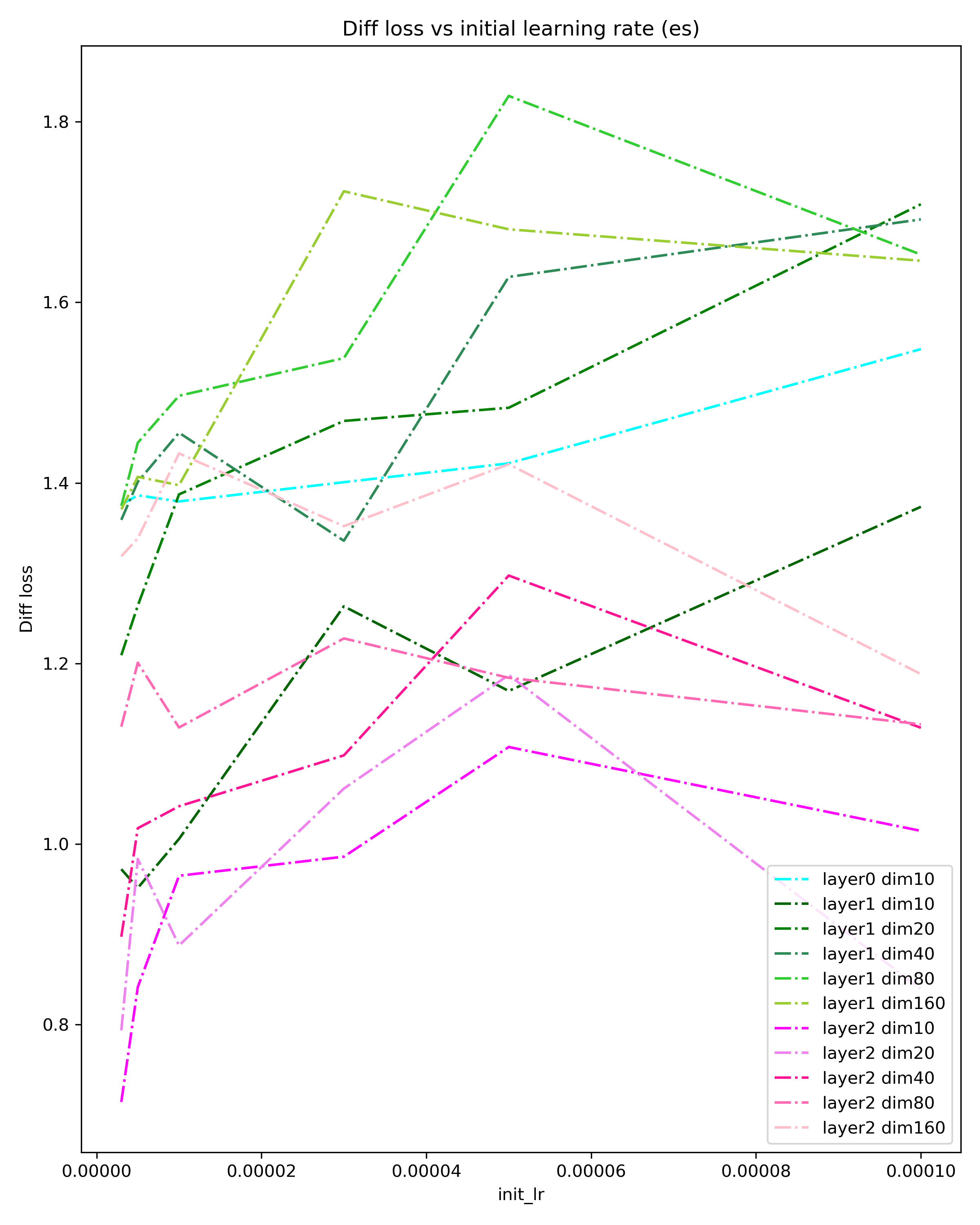}
    \caption{The ``difference of accuracy'' \citep{hewitt-liang-2019-designing} and the ``difference of loss'' \citep{pimentel2020information} criteria with different learning rates on model configurations, on UD Spanish. The weight decay is set to 0.}
    \label{fig:lr_vs_criteria_es}
\end{figure*}

\paragraph{Smaller probes are not necessarily better}
We find that while smaller probes have higher ``selectivity'' and ``information gain'' for mBERT representations, probes with one hidden layer and 40-80 hidden neurons are better than more simplistic probes, as shown in Figures \ref{fig:lr_vs_criteria_en}, \ref{fig:lr_vs_criteria_fr} and \ref{fig:lr_vs_criteria_es}. The plots show consistency between the two criteria. For example, larger models and more layers do not necessary lead to better results. Neither are the smallest probes with 0 hidden layers.

\begin{figure*}
    \centering
    \includegraphics[width=.49\linewidth]{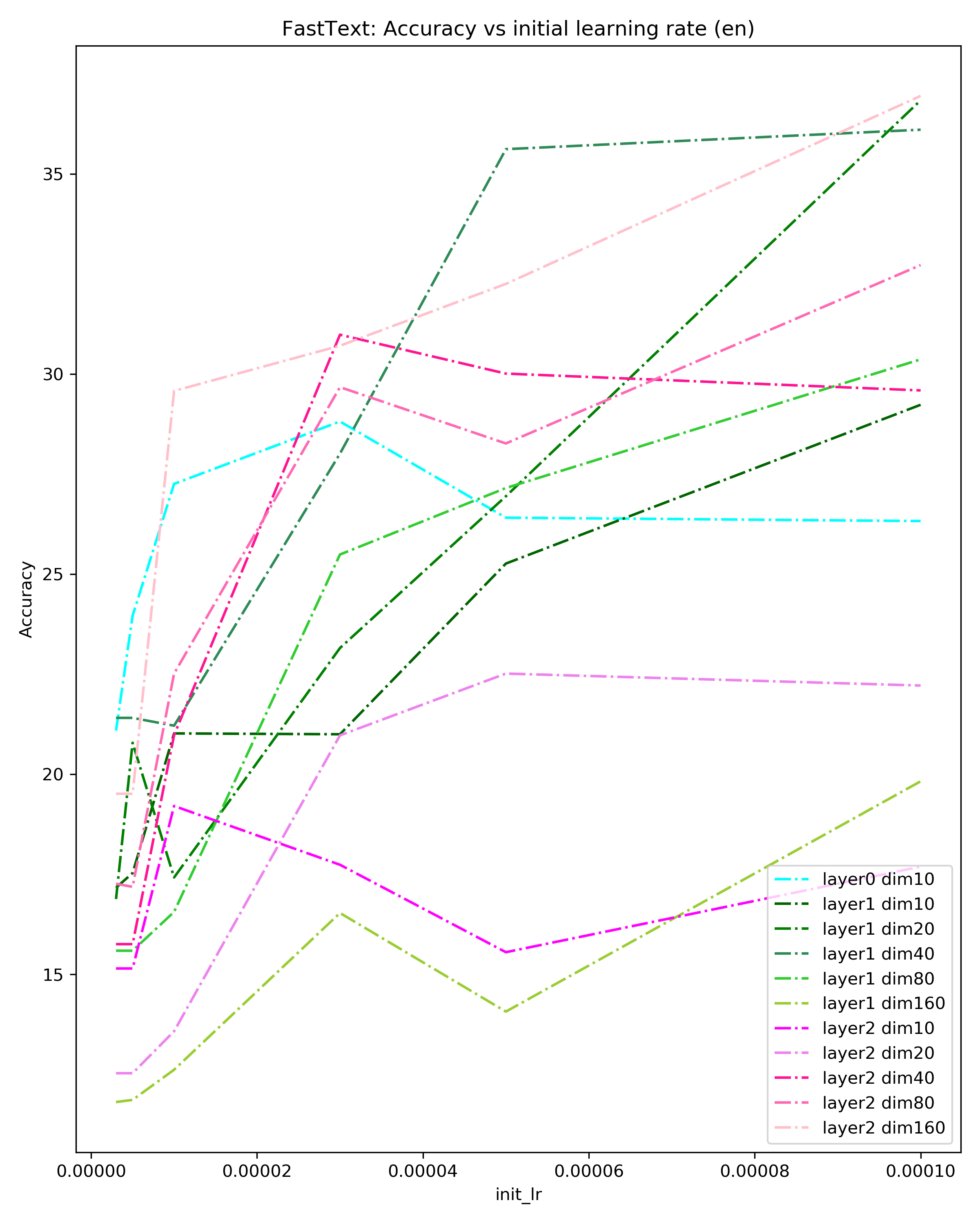}
    \includegraphics[width=.49\linewidth]{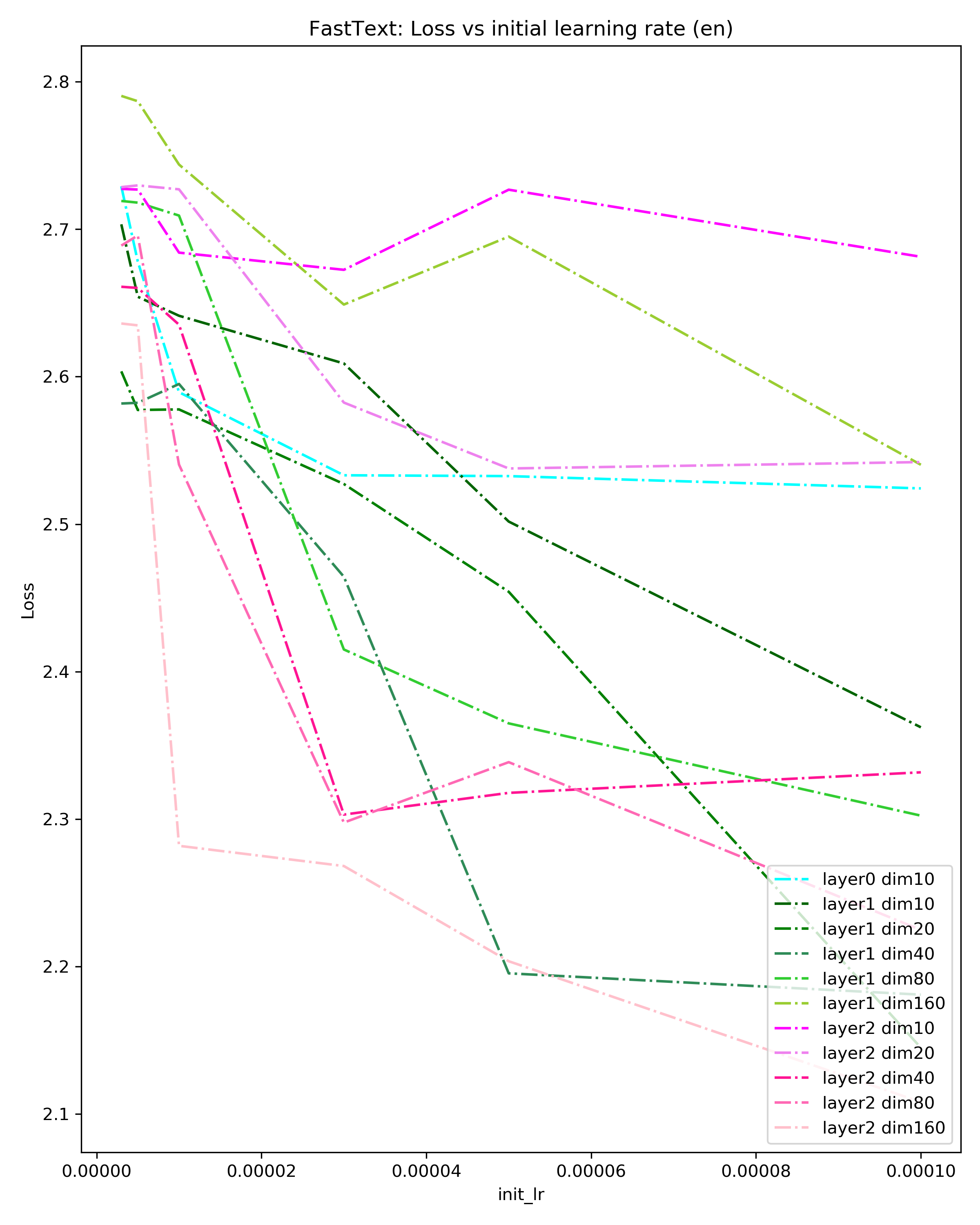}
    \caption{The accuracy and cross entropy loss of probes on FastText. These performances are much worse than those on mBERT, indicating the richness of information encoded in contextuality of mBERT.}
    \label{fig:fasttext_lr_acc_loss_en}
\end{figure*}

\begin{figure*}
    \centering
    \includegraphics[width=.49\linewidth]{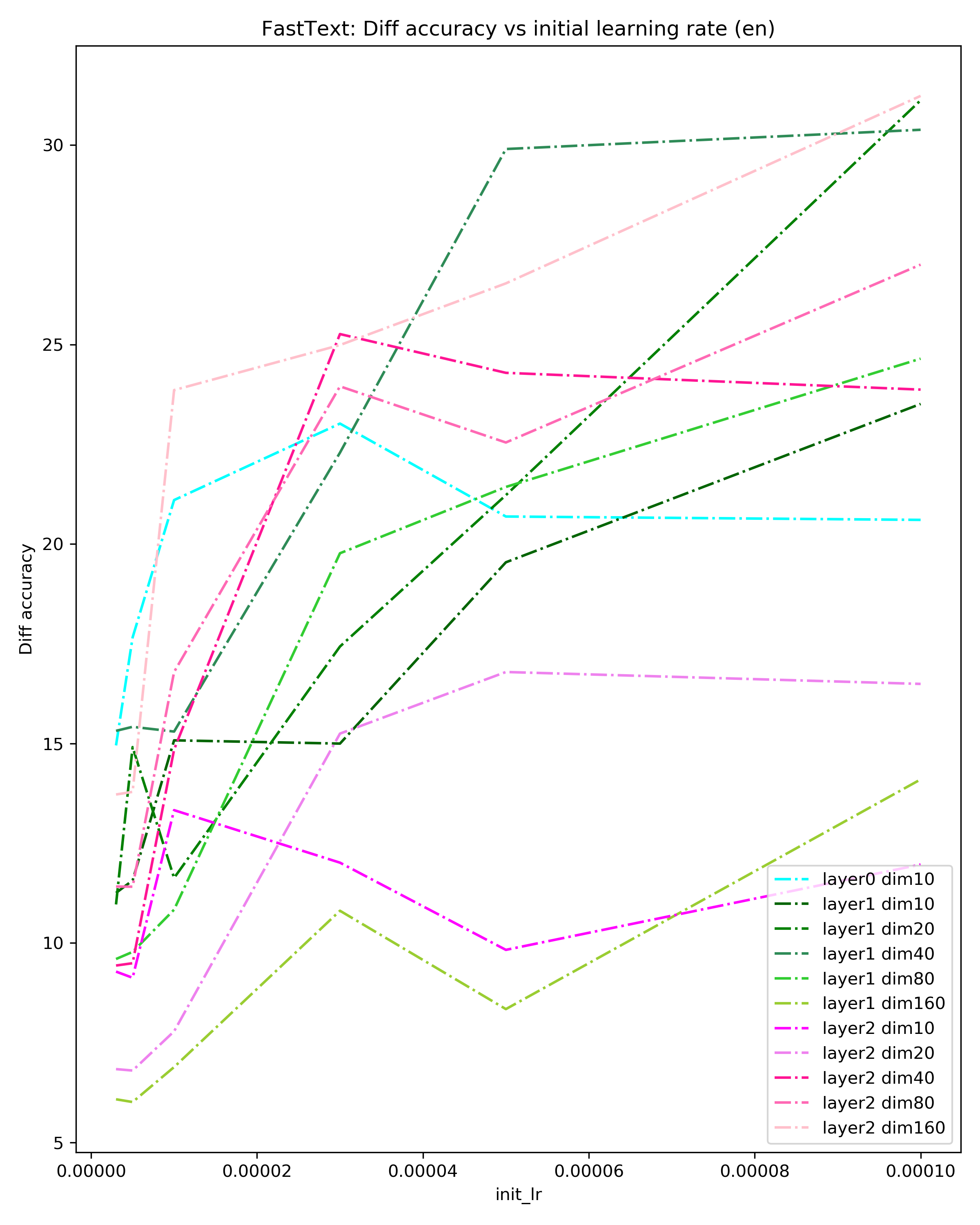}
    \includegraphics[width=.49\linewidth]{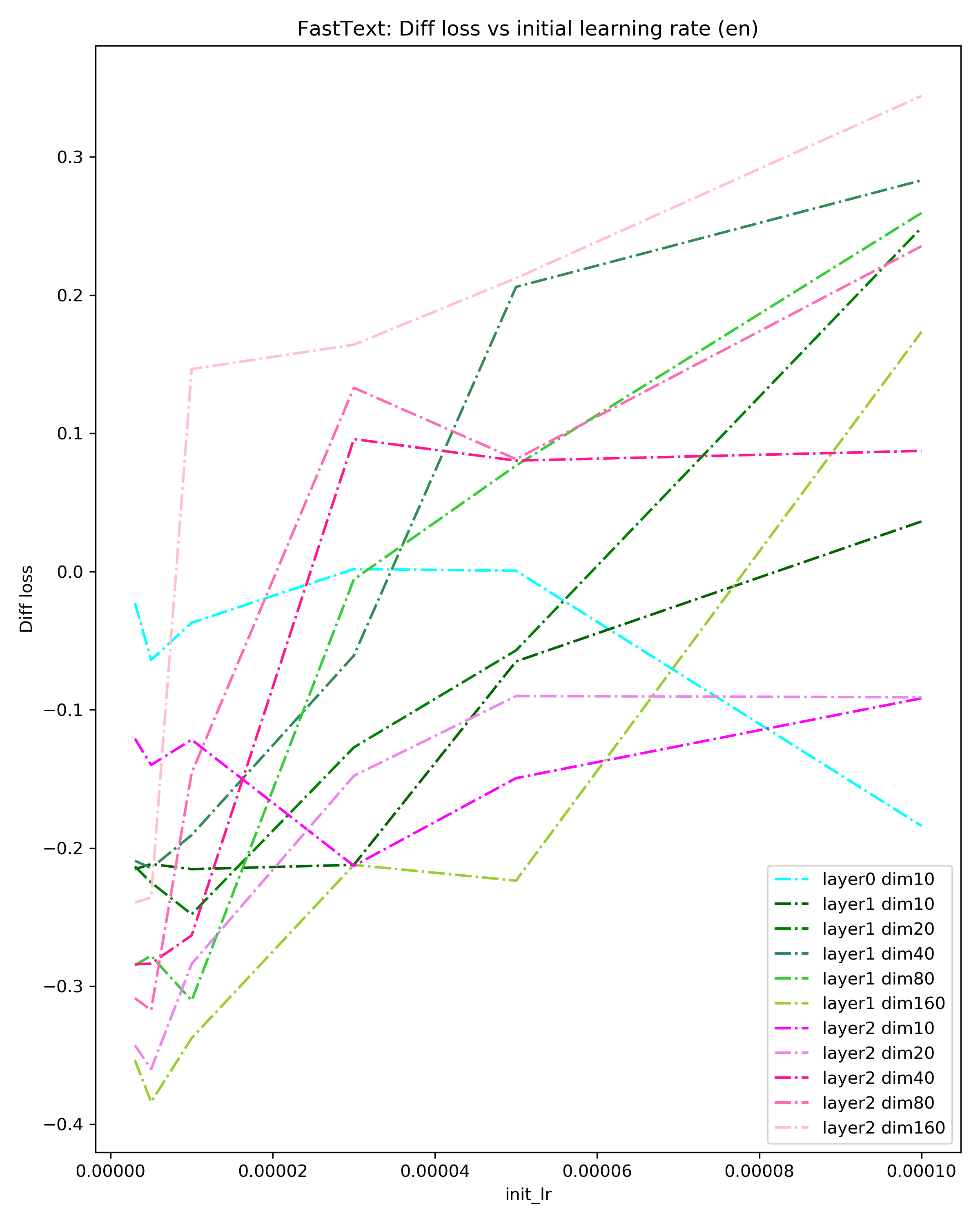}
    \caption{The selectivity \citep{hewitt-liang-2019-designing} and information gain \citep{pimentel2020information} of probes on FastText. Probes with different capacities are ranked similarly using these two criteria.}
    \label{fig:fasttext_criteria_en}
\end{figure*}

Note that we also swept hyperparameters for FastText, where probes with less parameters do not always outperform more complex probes in either accuracy, loss, selectivity, or information gain. Figures \ref{fig:fasttext_lr_acc_loss_en} and \ref{fig:fasttext_criteria_en} illustrate these observations.

\section{Reproducibility}
On a T4 GPU card, training one epoch takes around 20 seconds. Without setting maximum gradient steps, $98.6\%$ of experiments finish within 400 epochs. We will open source our codes.

%% Early Stopping plots
%%%%%%%%%%%%%%%%%%%%%%%%%%%

\iffalse  % These are not interesting. One is enough
\begin{figure*}
    \centering
    \includegraphics[width=\linewidth]{fig/maxgradstep_acc_fr.png}
    \includegraphics[width=\linewidth]{fig/maxgradstep_ent_fr.png}
    \caption{Max gradient step vs accuracy, t\_acc, f\_acc (blue) and loss, t\_ent, f\_ent (green) on French. The ``t'' refers to control task \citep{hewitt-liang-2019-designing}, and ``h'' refers to control function \citep{pimentel2020information}.}
    \label{fig:early_stopping_fr}
\end{figure*}
\begin{figure*}
    \centering
    \includegraphics[width=\linewidth]{fig/maxgradstep_acc_es.png}
    \includegraphics[width=\linewidth]{fig/maxgradstep_ent_es.png}
    \caption{Max gradient step vs accuracy, t\_acc, f\_acc (blue) and loss, t\_ent, f\_ent (green) on Spanish. The ``t'' refers to control task \citep{hewitt-liang-2019-designing}, and ``h'' refers to control function \citep{pimentel2020information}.}
    \label{fig:early_stopping_es}
\end{figure*}
\fi 

%% Weight Decay plots
%%%%%%%%%%%%%%%%%%%%%%%%%%%

%% Learning Rate plots
%%%%%%%%%%%%%%%%%%%%%%%%%%%

%% FastText learning rate plots
%%%%%%%%%%%%%%%%%%%%%%%%%%%%%%%%

\end{document}